\documentclass[runningheads,orivec]{llncs}
\usepackage[T1]{fontenc}
\usepackage{graphicx}
\usepackage[misc,geometry]{ifsym}

\usepackage{amsmath} 
\usepackage{amssymb}  
\usepackage{textcomp}
\usepackage{gensymb}
\usepackage{siunitx} 
\usepackage[table,xcdraw]{xcolor}
\usepackage{svg}
\usepackage{orcidlink}
\usepackage{hyperref}
\usepackage[english]{babel}
\addto\extrasenglish{

}
\graphicspath{ {./Images/} }

\usepackage{booktabs}

\begin{document}
\title{Graph-based Robot Localization Using a Graph Neural Network with a Floor Camera and a Feature Rich Industrial Floor}
\titlerunning{Graph-Based Robot Localization with GNNs on Industrial Floors}
%
\author{Dominik Brämer\textsuperscript{(\Letter)}\orcidlink{0009-0003-9326-432X}\inst{1}, Diana Kleingarn\orcidlink{0009-0001-1751-0504}\inst{1} \and Oliver Urbann\orcidlink{0000-0001-8596-9133}\inst{2}}
\authorrunning{D. Brämer, D. Kleingarn \and O. Urbann}
%
\institute{Robotics Research Institute, Section Information Technology,\\
TU Dortmund University, 44227 Dortmund, Germany \and
Fraunhofer IML,\\ Joseph-von-Fraunhofer-Str. 2-4, Dortmund, Germany\\
\email{dominik.braemer@tu-dortmund.de}}
\maketitle              

\begin{abstract}
Accurate localization represents a fundamental challenge in robotic navigation. 
Traditional methodologies, such as Lidar or QR-code-based systems, suffer from inherent scalability and adaptability constraints, particularly in complex environments. 
In this work, we propose an innovative localization framework that harnesses flooring characteristics by employing graph-based representations and Graph Convolutional Networks (GCNs). 
Our method uses graphs to represent floor features, which helps localize the robot more accurately (\SI{0.64}{\centi\metre} error) and more efficiently than comparing individual image features.
Additionally, this approach successfully addresses the kidnapped robot problem in every frame without requiring complex filtering processes. 
These advancements open up new possibilities for robotic navigation in diverse environments.

\end{abstract}
\section{Introduction}
\label{sec:intro}
The localization of autonomous mobile robots in indoor environments is a foundational challenge in robotics, driving innovations in sensor technologies and computational methods. 
Accurate position estimation is essential for effective navigation, task execution, and operational efficiency. 
While Global Positioning System (GPS) offers precise localization, it is unsuitable for indoor applications. 
Ultra-wideband (UWB) systems require costly hardware installations, and Light Detection and Ranging (Lidar) sensors, despite their popularity, vary significantly in cost and performance, creating barriers to widespread adoption.

Camera-based systems provide an alternative solution by utilizing artificial markers such as QR codes. 
However, QR-code-based approaches are both expensive and cumbersome, necessitating precise placement and regular maintenance of markers to ensure accurate localization. 
Forward- or ceiling-facing cameras~\cite{zhang2015} are often used but struggle with detecting smaller features due to occlusion and large distances, limiting their applicability.

Ground-facing cameras have been proposed to overcome these shortcomings by leveraging flooring patterns for localization~\cite{piet2024}. 
Although promising, this approach introduces new challenges, including computational inefficiencies, limited algorithmic runtime, and reduced adaptability. 
Studies indicate that algorithms tailored to specific floor types achieve better localization accuracy compared to generalized methods~\cite{zhang2019high}.

This paper presents a novel graph-based localization framework aimed at enhancing the capabilities of ground camera systems. 
By abstracting small features into graph representations, this approach reduces dependence on individual features while improving scalability and adaptability across diverse flooring types. 
Adjustments are straightforward, requiring modifications only to the detector component when necessary, making the proposed method both flexible and efficient.

\subsection{Related Work}
\label{related_work}
Ground cameras for localization have predominantly utilized SIFT or \mbox{CenSure} for feature extraction, as both are well-suited for this task~\cite{schmid2020features}. 
The hardware system by Kozak et al., employing \mbox{CenSure} and ORB for faster runtime and licensing reasons, achieves an accuracy of \SI{2}{\centi\metre} on asphalt but relies on initialization and previous positions~\cite{kozak2016ranger}.

Wang et al. developed a SLAM-like algorithm~\cite{WangPHY24} that integrates the Micro-GPS system by Zhang et al.~\cite{zhang2019high}, supporting diverse floor types such as asphalt, concrete, tiles, and carpet. 
Schmid et al.'s survey highlights the suitability of flooring features for localization but emphasizes that not all detector-descriptor combinations perform well on all floor types~\cite{schmid2020features}.

To address these limitations, Zhang et al. proposed a CNN autoencoder in 2018~\cite{zhang2018learning}, enhancing floor feature descriptors but limiting testing by reidentifying textured floor images without ground truth evaluations. 
Another unpublished framework builds on this concept, combining a CNN autoencoder with ground truth evaluation, achieving a position error of \SI{2}{\centi\metre} and a rotation error of \SI{2.4}{\degree}~\cite{piet2024}.

Most ground camera systems share a common structure comprising three stages:
\begin{itemize}
    \item[] Stage 1: Feature detection and description.
    \item[] Stage 2: Database search.
    \item[] Stage 3: Localization.
\end{itemize}

\subsection{Contribution}
Existing applications face challenges in scalability and adaptability, with the search space expanding rapidly as the number of ground features increases. This growth makes Stage 2 a bottleneck for many systems, despite efforts like PCA-based descriptor optimization, which still results in some performance loss~\cite{piet2024,zhang2019high}. Existing applications face multiple issues this include:
\begin{itemize}
    \item Scalability limitations due to the rapid growth of features.
    \item Inaccurate localization assessments without real ground truth data.
    \item Difficulty adapting to different floor characteristics.
\end{itemize}
This work addresses these issues by leveraging graphs combined with machine learning:
\begin{itemize}
    \item Comparing graphs instead of individual features enhances scalability.
    \item Graphs provide abstraction layers that improve adaptability.
    \item High precision ground truth data from a Vicon system enables robust evaluations.
\end{itemize}
This novel graph-based solution focuses on increasing scalability and adaptability to overcome the limitations of existing methods.

\begin{figure}[htbp]
    \centering
    \begin{minipage}[b]{0.45\textwidth}
    \centering
    \includegraphics[width=0.9\textwidth]{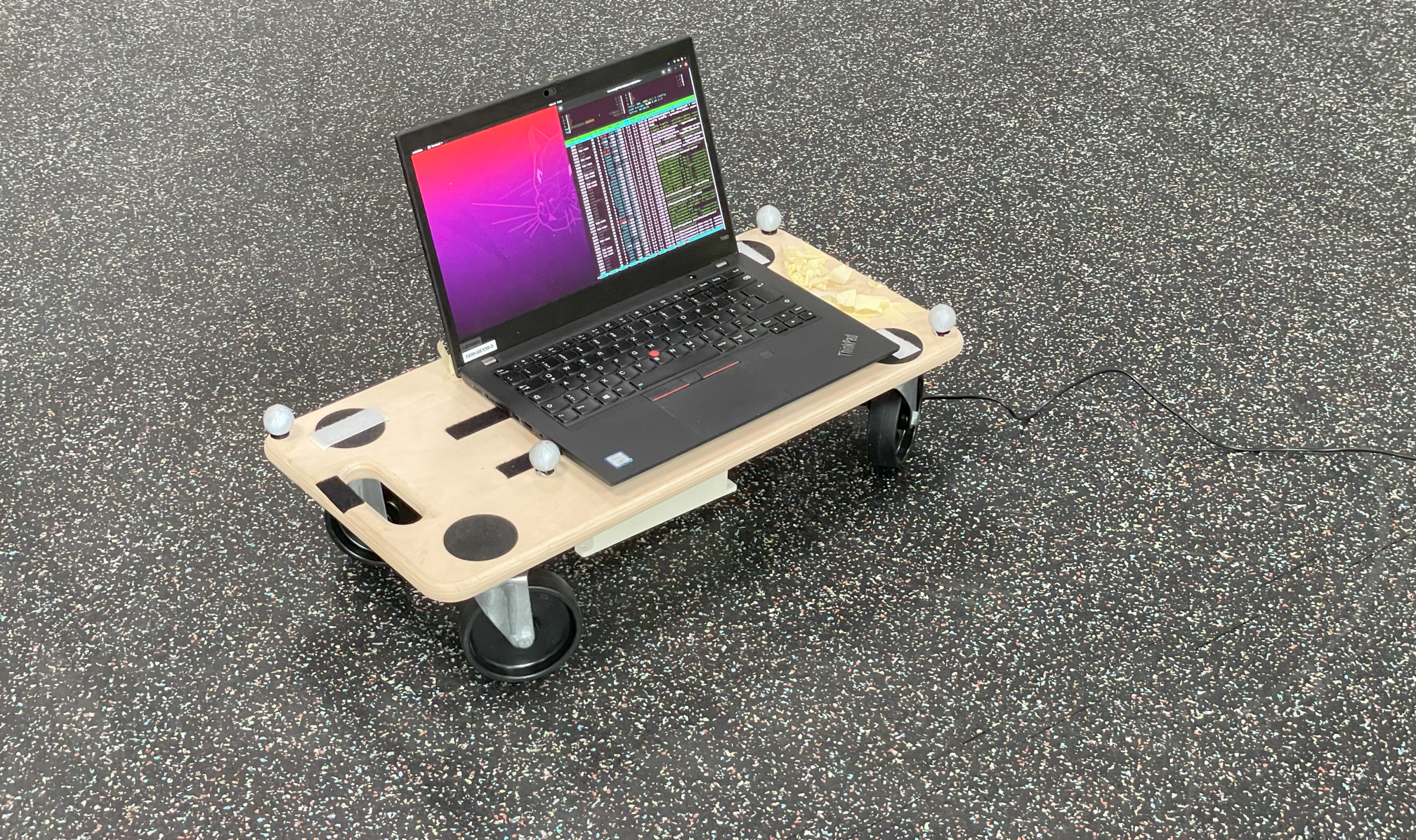} 
    \caption{The robot dummy was used to gather the required images in combination with their corresponding locations. The motion-capturing system tracks the robot dummy using spherical silver markers.}
    \label{fig:robot}\end{minipage}
    \hfill
    \begin{minipage}[b]{0.45\textwidth}
    \centering
    \includegraphics[width=0.9\textwidth]{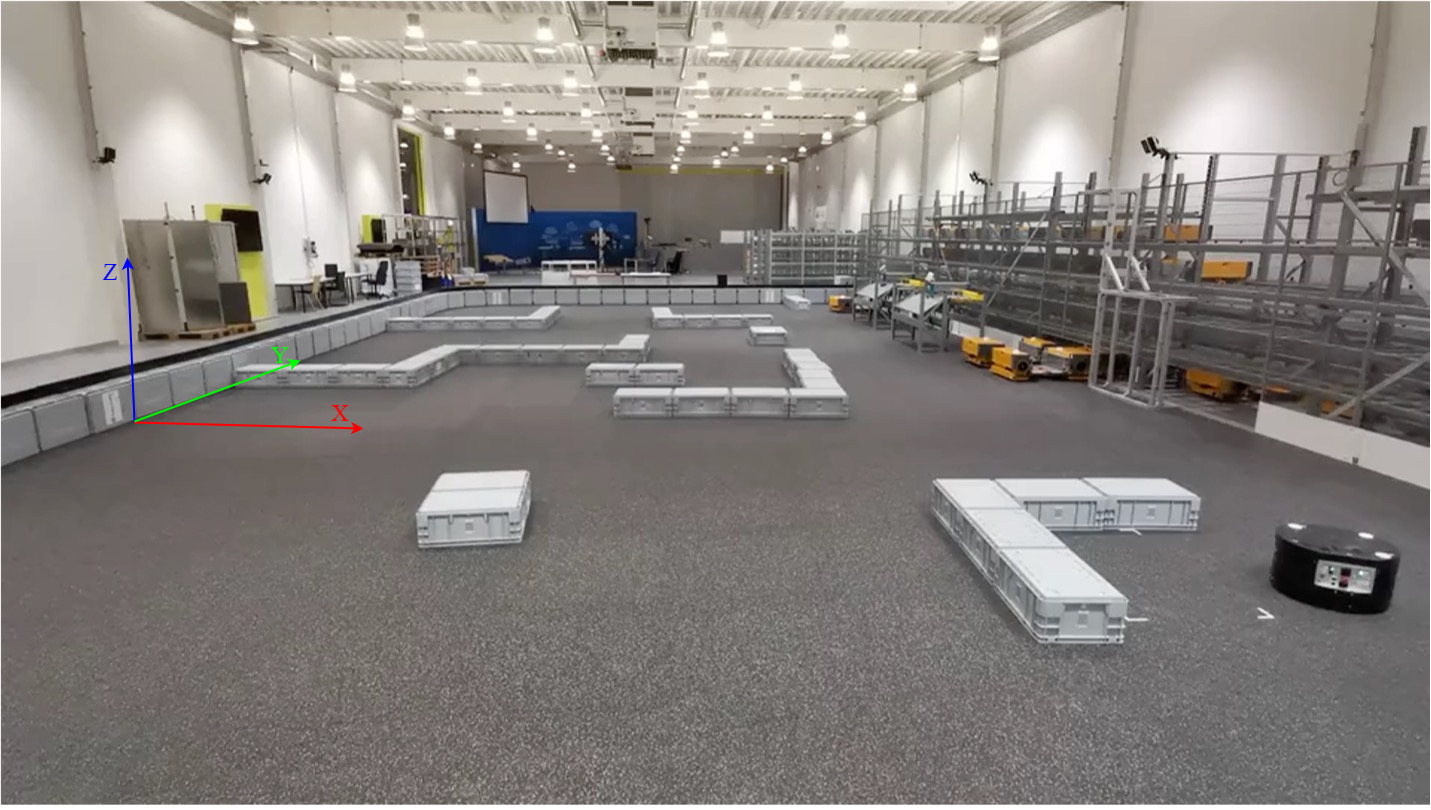} 
    \caption{The logistic hall at Fraunhofer IML displays a color-coded coordinate system representing the world coordinate system of the Vicon motion capturing system. It is shifted to the side for better visibility.}
    \label{fig:hall}
    \end{minipage}
\end{figure}

\section{Dataset}
\label{dataset}
Our dataset offers a key advantage with real ground truth positioning data obtained via a Vicon motion capture system. 
While Zhang et al. provide a larger dataset~\cite{zhang2019high} covering various floor types\footnote{https://microgps.cs.princeton.edu/~\cite{zhang2019high}}, their dataset relies on estimated positions from image stitching. 
In our approach, the Vicon system tracks our marker-equipped dummy robot (see \autoref{fig:robot} and \autoref{fig:hall}), allowing simultaneous capture of floor images at \SI{60}{\hertz} and position data at \SI{200}{\hertz}. 
Images are recorded using the UI-3131LE camera (IDS GmbH), which supports up to \SI{135}{\hertz} at a resolution of $808 \times 608$ pixels with a global shutter, covering an area of $10.9 \times 8.2$\,cm per image. 
Each image is synchronized with its corresponding position, and the dataset is split into training, validation, and evaluation sets. 

Since we focus on exploring graph neural networks for improved generalization rather than developing a new localization framework, we use a small subset covering one square meter combined with parts of Zhang et al.'s dataset for evaluation. 
Note that the ground truth data has been adjusted due to motion detection and synchronization errors, resulting in a position accuracy of \SI{0.5}{\centi\metre}.

\section{System Design}
\label{sec:sysdes}
In this Section, we introduce the graph neural network, which combines multiple detected features of a ground image of the floor into a single graph and computes a description for this combined feature \mbox{(Stage 1)}.
We also present the mapping process used for database creation \mbox{(Stage 2)} and the position estimation process \mbox{(Stage 3)}.

\subsection{Graph Design}
\label{sec:graphDesign}
For our approach, we use undirected graphs, denoted as $G=(V, E)$, where $V$ represents the set of nodes and $E$ represents the set of edges with $E \subseteq V \times V$. Each node $\mathit{v} \in V$ is defined as a vector $\mathit{v} = (m, n)$, representing the pixel position of the node within the ground image $\mathbf{I}$; this vector is also referred to as the node feature. The set of edges $E$ consists of node pairs $(\mathit{v}_i,\mathit{v}_j) \in E$ with $\mathit{v}_i,\mathit{v}_j \in V$, and due to the graph's undirected nature, $(\mathit{v}_j,\mathit{v}_i) \in E$ as well.

The crosswise distances between every node, measured in pixel, define the edge weight of the graph, and the adjacency matrix is denoted as $\mathbf{A} \in {\mathbb{R}_0^{+}}^{|V| \times |V|}$. 
This is necessary to compute a better modified Laplacian matrix for the later steps.
For any edge $(\mathit{v}_i, \mathit{v}_j) \in E$ from node $\mathit{v}_i$ to node $\mathit{v}_j$, we have $\mathbf{A}_{\mathit{v}_i,\mathit{v}_j} = w$, where $w \in \mathbb{R}_0^{+}$ represents the weight between node $\mathit{v}_i$ and node $\mathit{v}_j$.
Furthermore, each node $\mathit{v} \in V$ is associated with a SIFT keypoint description $\mathit{k} \in \mathbb{R}^{128}$ through the mapping $f_{map}: V \rightarrow \mathbb{R}^{128}, \mathit{v} \mapsto \mathit{k}$.

This graph-forming process allows us to combine all the information generated from an image in a graph $G$ with the additional details of the adjacency matrix $\mathbf{A}$ and the SIFT keypoint description mapping.

\subsection{Graph Convolutional  Network}
\label{sec:GNN}
The proposed architecture of the Graph Convolutional Network (GCN) is inspired by the SimGNN network~\cite{bai2019simgnn} and focuses on graph-level embedding, compactly representing a graph and its structure. 
Our graph-level embedding is built using three GCN Convolution Layers~\cite{kipf2016semi} and a Global Attention Sum Pool Layer, transforming a graph $G$  into a compact description.

We utilize a Siamese Neural Network (SNN)~\cite{chicco2021siamese} framework to train our graph encoder, which splits into three components: the encoder, the distance calculation, and the loss function. 
The GCN of the encoder part takes the following form:
\begin{align}
\label{eq:gcn_math_example}
f_{enc}(\mathbf{X}, \hat{\mathbf{A}}) = \sum^{N}_{i=0} \alpha_i \cdot \hat{\mathbf{A}}\left( \hat{\mathbf{A}} \left( \hat{\mathbf{A}} \mathbf{X} \mathbf{W}^{0} \right) \mathbf{W}^{1}\right) \mathbf{W}^{2}\text{~\cite{kipf2016semi}},
\end{align}
which describes the GCN proposed by Kipf et al.
Here, the adjacency matrix $\mathbf{A}$ of an undirected graph $G$ together with an identity matrix $\mathbf{I}_N$ builds $\Tilde{\mathbf{A}} = \mathbf{A} + \mathbf{I}_N$~\cite{kipf2016semi}.
The trainable and layer-specific weight matrices are formed by $\Tilde{\mathbf{D}}_{ii} = \sum_j \Tilde{\mathbf{A}}_ij$ and $\mathbf{W}^{(l)}$, where $l$ denotes the l-th layer~\cite{kipf2016semi}.
So that $\hat{\mathbf{A}} =  \Tilde{\mathbf{D}}^{-\frac{1}{2}} \Tilde{\mathbf{A}} \Tilde{\mathbf{D}}^{-\frac{1}{2}}$, and $\mathbf{X}$ is the resulting matrix when $f_{map}$ is applied for every node $v \in \mathbf{V}$ of a graph $G$.
We apply the softmax function~\cite{kipf2016semi} that is applied in a row-wise manner as an activation function for the resulting embedding.

In the distance part, we compute the Euclidean distance between the embeddings of two graphs to obtain a similarity score. 
For the loss function, we use the Contrastive-Loss~\cite{khosla2020supervised}, as proposed by Le Cunn et al.~\cite{chopra2005learning}, which is linked to metric distance learning or triplets~\cite{khosla2020supervised,chopra2005learning}.

This work explores how curriculum learning (CL) affects result accuracy. 
Instead of shuffling training data randomly, CL organizes it by difficulty, from easy to hard to learn~\cite{soviany2022curriculum}. 
We will focus on Data-Level CL, which is also a straightforward realization of the general concept of curriculum learning.

\subsection{Mapping}
\label{sec:mapping}
In \autoref{sec:graphDesign} and \autoref{sec:GNN}, we introduced our GCN, which distinguishes similar and dissimilar graphs as the first component of our localization method.
The second component is a floor map, defined by the mapping $f_{pos}: \mathbb{I} \rightarrow \mathbb{R}^{3}, \mathit{(m, n)} \mapsto \mathit{p}$ where $(m,n)$ is a pixel in an image $\mathbf{I} \in \mathbb{I}$, $x$ and $y$ 
 are absolute world coordinates on a projected floor plane (see \autoref{fig:hall}), and $r$ is the rotation about the z-axis in degrees.

We generate this mapping by moving a dummy robot along a zig-zag pattern while recording images and positions in parallel (as illustrated in \autoref{fig:zig_zag} and \autoref{fig:vicon_cameras}).
Due to slight inaccuracies in the motion capture system and synchronization errors, our ground truth data contains minor errors, further detailed in \autoref{sec:eval}.
This combined dataset—associating each image with a world position—enables mapping image-based graphs to real-world positions. 
Using our GCN, described in \autoref{sec:GNN}, we build a Map-Database that links graph embeddings to their corresponding positions and rotations.

\subsection{Position Estimation}
\label{sec:positionEstimation}
In \autoref{sec:graphDesign}, \autoref{sec:GNN}, and \autoref{sec:mapping}introduced our image-to-position framework, forming the basis of our position estimation pipeline.
First, an image $I$ is captured and processed using the SIFT algorithm to extract 128 features, creating a graph $G = (V,E)$ with $|V| = 128$ and a corresponding adjacency matrix $\mathbf{A}_{128 \times 128}$ representing pairwise distances. 
After preprocessing $\mathbf{A}$ into $\hat{\mathbf{A}}$ and mapping the nodes via $f_{map}$ to form the feature matrix $\mathbf{X}$, the GCN, described in \autoref{sec:graphDesign}, computes an embedding vector $e$. 
We then retrieve the n nearest embeddings from the database built in \autoref{sec:mapping} and use their position data to predict the current position, reducing the need for repeated GCN executions by relying on a simpler distance function.

Every graph $G$ includes $V$, and every node $v \in V$ is composed of its location inside of the image $\mathbf{I}$ it belongs to.
This additional information is stored in a separate Graph-Database, while the mapping data forms the Map-Database. 
Together, these datasets enable us to associate graph embeddings with their corresponding world positions.

Next, we project these feature positions into world coordinates by estimating a homography matrix~\cite{dubrofsky2009homography} using positions from the Graph-Database ($P_{base}$) and those of the generated graph ($P_{local}$). 
This homography yields one rotation candidate per nearest graph. 
We use the median of these n candidates as the current rotation, to filter out occasionally occurring outliers. 
To further reduce position errors incurred from considering many nearest embeddings, we apply the DBSCAN~\cite{schubert2017dbscan} clustering algorithm on the retrieved positions, and the mean of the largest cluster is used as the final position estimate.
This ensures that even if the procedure is carried out with a larger number of neighbors and thus possible outliers, a cluster point and thus the presumed position can be found.

\section{Evaluation}
\label{sec:eval}
The test setup used for the evaluation consists of several components, all of them are presented and explained below.
Furthermore, the metrics used for the evaluation are explained in the following sections.

\begin{figure}[htbp]
    \centering
    \begin{minipage}[b]{0.45\textwidth}
        \centering
        \includegraphics[width=\textwidth]{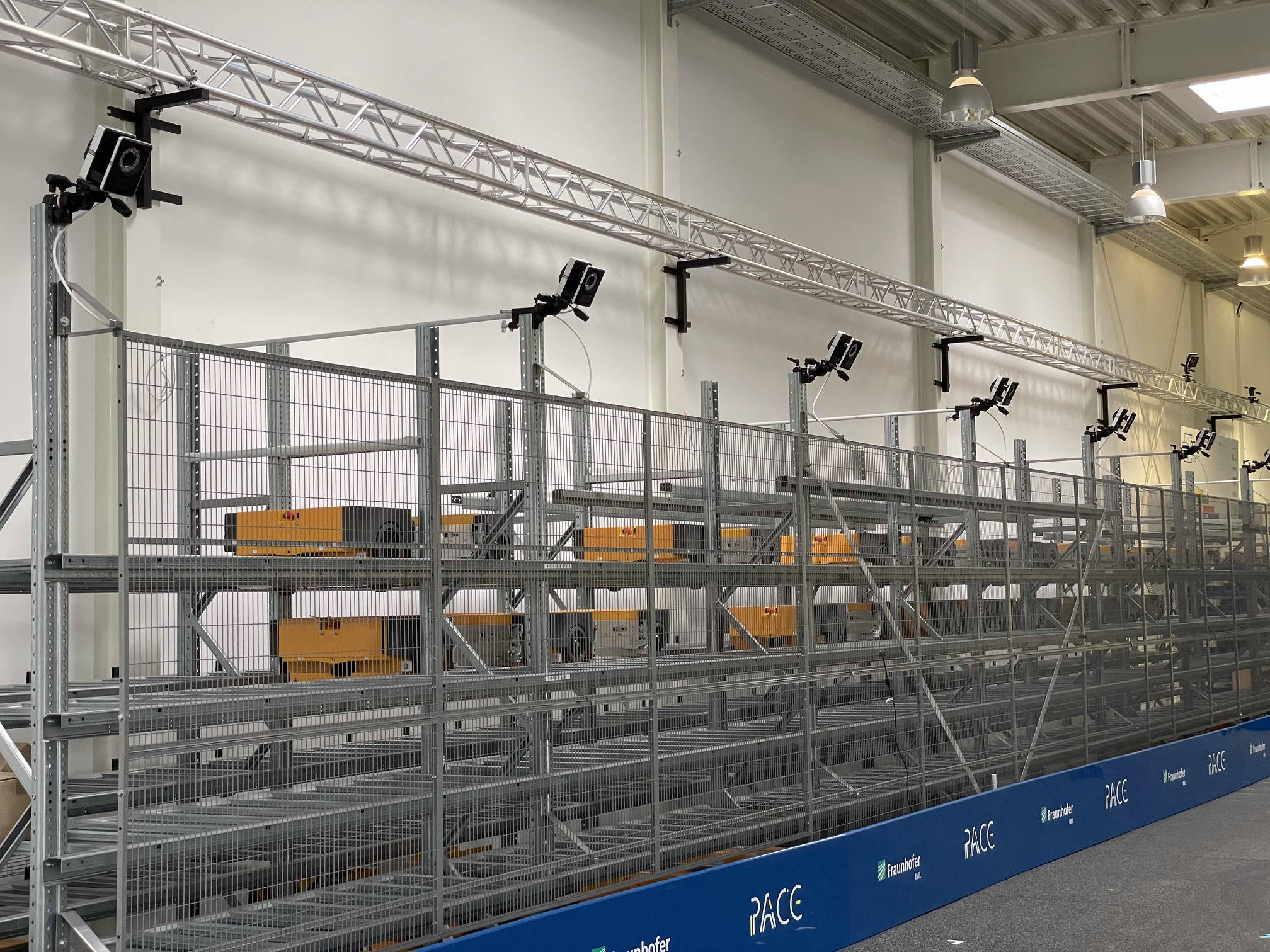}
        \caption{Illustration of the cameras of the motion capturing system from the company Vicon.}
        \label{fig:vicon_cameras}
    \end{minipage}
    \hfill
    \begin{minipage}[b]{0.5\textwidth}
        \centering
        \includegraphics[width=\textwidth]{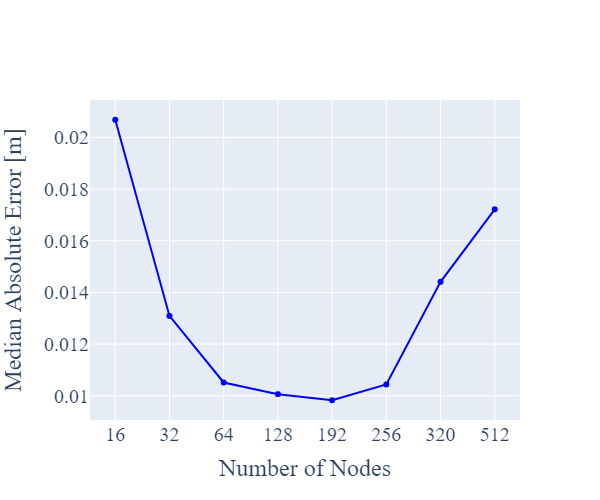}
        \caption{Effect of graph node count (used in GCN training) on localization error (m); lower is better.} 
        \label{fig:node_position_error}
    \end{minipage}
\end{figure}

In addition to the created dataset, we also use existing datasets\footnote{https://microgps.cs.princeton.edu} of various floor types created by Zhang et al.~\cite{zhang2019high}.
Since these datasets do not provide real ground truth data, we will use a different metric to evaluate this data than to evaluate the self-generated data.

Our dataset is evaluated by the median absolute localization error in meters and the median absolute rotation error in degrees. The additional floor dataset is assessed using a best-out-of-10 approach based on pixel distance between the images, calculated from the data used to assemble a stitched image from individual images in the dataset.

As a reference for evaluation, we employ a naive graph similarity measure that computes the mean minimum distance between node features of two graphs. 
This approach deliberately disregards the underlying graph structure, resulting in a localization method that is conceptually closer to the floor localization techniques introduced by Zhang et al. and Brömmel et al. \cite{zhang2019high,piet2024}.
For simplicity, we will refer to this method as the baseline throughout the paper.

The training and evaluation are done on a Lenovo X1 Extreme notebook using an Intel Core i7-8750H CPU and an NVIDIA GeForce GTX 1050Ti with Max-Q graphics card.

\begin{figure}[htbp]
    \centering
    \includegraphics[width=0.45\textwidth]{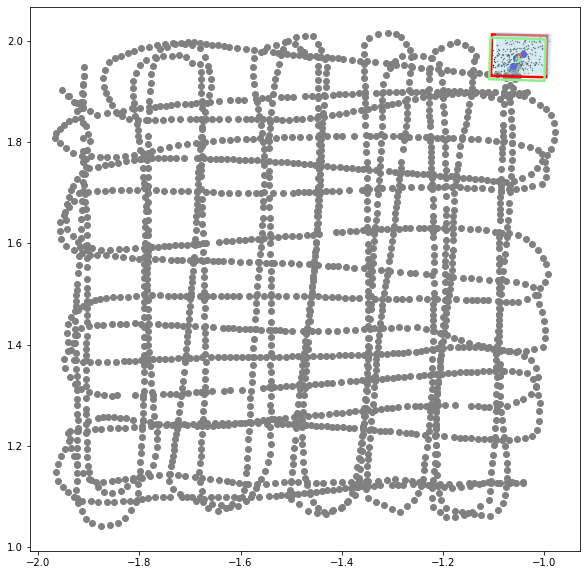}
    \caption{The course was recorded in the experimental logistics hall. Grey dots indicate images/graphs in the Graph-Database; large blue dots represent graphs used for position estimation, while small blue dots show projected feature locations. The boxes depict the robot dummy’s field of view: the green box (centered at the green dot) indicates the ground truth, the light blue box marks the area of the neighbors used for the prediction, and the red box (centered at the red dot) shows the estimated position.}
    \label{fig:zig_zag}
\end{figure}

\begin{figure}[htbp]
    \centering
    \begin{minipage}[b]{0.45\textwidth}
    \includegraphics[width=1\textwidth]{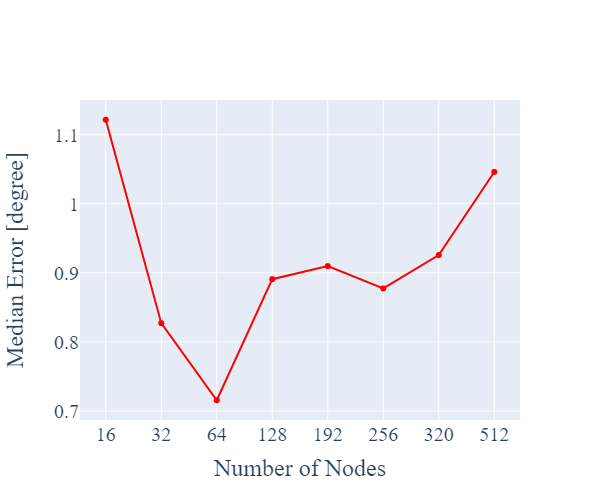}
    \caption{Effect of graph node count (used in GCN training) on rotation error in degrees for our localization pipeline; lower is better.}
    \label{fig:node_rotation_error}
    \end{minipage}
    \hfill
    \begin{minipage}[b]{0.45\textwidth}
    \includegraphics[width=1\textwidth]{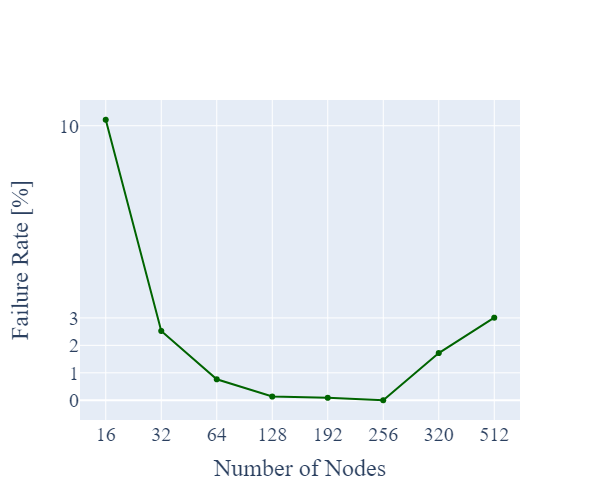}
    \caption{Failure rate of rotation refinement vs. graph node count (nodes used in GCN training); lower is better.}
    \label{fig:node_failure_rate}
    \end{minipage}
\end{figure}

We trained our GCN with graphs of different sizes to analyze the impact of node quantities on position accuracy. Then, we ran a test where a path was traversed twice on a one square meter area in a horizontally and vertically zig-zag course (\autoref{fig:zig_zag}).
To obtain sufficient coverage of this area, 2232 images were required.
The analyzed node quantities are 16, 32, 64, 128, 192, 256, 320, and 512. 
Each network is trained with graphs generated from a single node quantity, undergoing a maximum of 1000 training epochs with early stopping. 
We use the median absolute rotation error and a failure rate percentage to indicate cases where rotation estimation was not possible. 
Additionally, we use the 15 nearest embeddings and graphs for position and rotation prediction.
\autoref{fig:node_position_error} shows that the number of nodes of a graph influences the position error of our localization pipeline.
It also shows that the accuracy decreases rapidly for less than 64 nodes and decreases rapidly for more than 256.
In addition, the accuracy of the position for graphs with a number of nodes between 64 and 256 at a similar level, with a sweet spot at 192 nodes.
The exact values can be found in \autoref{tab:accuracy}.
All tables in this evaluation will highlight the lowest possible values of our method, which also have the lowest failure rate.
\begin{table}[htbp]
\caption{The accuracies of the position and rotation as well as the failure rate of the rotation refinement; lower is better. The number of nodes indicates the graph nodes used to train the GCN.}
\vspace{-1px}
\centering
\resizebox{1\textwidth}{!} {
\begin{tabular}{@{}rccc@{}}
\toprule
\multicolumn{1}{c}{} & Position Error {[}m{]} & Rotation Error {[}degree{]} & Failure Rate {[}\%{]} \\ \midrule
16 Nodes  & 0.0207          & 1.1218          & 10.22        \\
32 Nodes  & 0.0131          & 0.827           & 2.52         \\
64 Nodes  & 0.0105          & 0.7154          & 0.76         \\
128 Nodes & 0.0101          & 0.8908          & 0.13         \\
192 Nodes & \textbf{0.0098} & 0.9097          & 0.09         \\
256 Nodes & 0.0104          & \textbf{0.8774} & \textbf{0.0} \\
320 Nodes & 0.0144          & 0.9257          & 1.72         \\
512 Nodes & 0.0172          & 1.0462          & 3.01         \\ \bottomrule
\end{tabular}
}
\label{tab:accuracy}
\end{table}
\autoref{fig:node_rotation_error} shows that the number of nodes of a graph also influences the rotation error of our localization pipeline.
The error of rotation is similar to the error of position, with the difference that a particularly small error of 0.7154 degrees is achieved for 64 nodes.
However, this can be explained by graphs with 64 nodes having a higher failure rate of nearly one percent than those with 128, 192, or 256 nodes.
This can also be seen in \autoref{fig:node_failure_rate}, which visualizes the failure rate for the different numbers of nodes.
There, it can also be observed that graphs with more than 256 nodes also have a higher failure rate, whereas graphs with a number of nodes between 128 and 256 stand out due to a shallow failure rate.

Graphs containing fewer than 128 nodes may not have sufficient intersections to accurately estimate the homography matrix. Conversely, graphs with more than 256 nodes may introduce excessive noise, making the estimation process more complex. Therefore, graphs with 64 to 256 nodes exhibit low errors and a minimal failure rate, and therefore warrant further evaluation.

To examine the impact of curriculum learning on position accuracy, GCNs are trained with data categorized by difficulty levels (easy, medium, and hard). Using a reduced number of maximum epochs prevents overfitting, as training converges faster.

The training data for the easy difficulty level consists of graphs generated from similar images and dissimilar images in terms of their world coordinates.
The medium difficulty level training data consists of graphs generated from one randomly selected image out of the five closest images in world coordinates, to ensure similarity.
In case the graphs should be dissimilar, the training data is composed of graphs where each graph is generated from a randomly selected image from the 10th to 100th furthest images.
The hard difficulty level uses training data similar to the medium difficulty level but excludes the two closest images for similar graphs. For dissimilar graphs, the training data is composed of graphs generated from randomly picked images between the 100th and 1000th furthest.

\begin{table}[htbp]
\caption{The accuracies of the position and rotation as well as the failure rate of the rotation refinement, influenced by curriculum learning, lower is better. The number of nodes indicates the number of nodes of the graphs used to train the GCN.}
\centering
\resizebox{1\textwidth}{!} {
\begin{tabular}{@{}rccc@{}}
\hline
\multicolumn{1}{c}{} & Position Error {[}m{]} & Rotation Error {[}degree{]} & Failure Rate {[}\%{]} \\ \hline
64 Nodes  & 0.0105        & 0.8216          & 0.22         \\
128 Nodes & \textbf{0.01} & 0.9647          & \textbf{0.0} \\
192 Nodes & \textbf{0.01} & 0.9463          & \textbf{0.0} \\
256 Nodes & \textbf{0.01} & \textbf{0.8671} & \textbf{0.0}
\end{tabular}
}
\label{tab:cl_accuracy}
\end{table}
\autoref{tab:cl_accuracy} presents the accuracies of the localization pipeline using GCNs trained on graphs consisting of 64, 128, 192, and 256 nodes. 
Notably, the localization pipeline trained with graphs of 64 nodes exhibits the largest position error, albeit only marginally, while achieving the smallest rotation error. 
This discrepancy is likely due to the inability to estimate the homography matrix in certain cases, where not enough overlapping keypoints can be found.
This results in fewer rotation estimates being included in the error calculation. 

For graphs with 128, 192, and 256 nodes used during training and evaluation, a clear trend emerges: while position estimation remains consistent, rotation estimation steadily improves. 
Furthermore, the localization pipeline achieves a zero failure rate across all these cases.
\begin{table}[htbp]
\caption{The accuracies of the position and rotation as well as the failure rate of the rotation refinement, influenced by curriculum learning. Lower is better. Here, the number of nodes stands for the number of graph nodes in the Map-Database. The GCN, used for all evaluations here, is trained exclusively with graphs with 256 nodes and the curriculum learning strategy.}
\centering
\resizebox{1\textwidth}{!} {
\begin{tabular}{rccc}
\hline
\multicolumn{1}{c}{} & Position Error {[}m{]} & Rotation Error {[}degree{]} & Failure Rate {[}\%{]} \\ \hline
16 Nodes  & 0.0139          & 1.0935          & 5.31         \\
32 Nodes  & 0.0109          & \textbf{0.8582} & 0.72         \\
64 Nodes  & 0.0102          & 0.9701          & 0.09         \\
128 Nodes & 0.0101          & 0.9576          & 0.04         \\
192 Nodes & \textbf{0.0099} & 0.9739          & \textbf{0.0} \\
256 Nodes & 0.01            & 0.8671          & \textbf{0.0} \\
320 Nodes & 0.0101          & 0.9894          & \textbf{0.0} \\
512 Nodes & 0.0102          & 1.0287          & \textbf{0.0} \\ \hline
\end{tabular}
}
\label{tab:256_cl_accuracy}
\end{table}
The localization pipeline, specifically the GCN trained exclusively on 256-node graphs, was further analyzed to assess its interaction with graphs containing fewer or more than 256 nodes. 
\autoref{tab:256_cl_accuracy} provides a detailed overview of the accuracies obtained during this investigation. 
The results indicate that position error, rotation error, and failure rate improve consistently with varying node counts, with 256 nodes demonstrating the optimal balance between accuracy and failure rate. 
Consequently, only the GCN trained and evaluated with 256 nodes will be used in subsequent evaluations of the localization pipeline.

Earlier evaluations determined the accuracy of position and rotation estimation based on the 15 nearest embeddings and graphs in the Map-Database. 
In this study, we further examine how the number of graphs impacts the accuracy of these estimations.

Since localization techniques are only practical if they operate within reasonable time constraints, the runtime performance of the pipeline is also analyzed in detail.
\begin{table}[htbp]
\caption{The accuracies of the position and rotation as well as the average runtime from graph to position, lower is better.
Here, the number of neighbors stands for the number of graphs used for the position estimation and refinement. The GCN, used for all evaluations here, is trained exclusively with graphs with 256 nodes and the curriculum learning strategy.}
\centering
\resizebox{1\textwidth}{!} {
\begin{tabular}{lccc}
\hline
\multicolumn{1}{c}{}                                    & Position Error {[}m{]} & Rotation Error {[}degree{]} & \multicolumn{1}{l}{Runtime {[}ms{]}} \\ \hline
\multicolumn{1}{c}{Baseline}                            & 0.0055 & 0.3376 & 3110.6 \\
\multicolumn{1}{r}{3 Neighbors} & \textbf{0.0064}        & \textbf{0.3944}             & \textbf{75.93}                       \\
\multicolumn{1}{r}{7 Neighbors} & 0.0086 & 0.6985 & 80.66   \\
\multicolumn{1}{r}{9 Neighbors} & 0.009  & 0.7534 & 85.11   \\
11 Neighbors                                            & 0.0094 & 0.8043 & 87.96   \\
15 Neighbors                                            & 0.01   & 0.8671 & 90.58   \\
19 Neighbors                                            & 0.0108 & 1.0237 & 92.32  
\end{tabular}
}
\label{tab:refinement_k_accuracy}
\end{table}
The graphs utilized for position estimation are the sole focus of this study, with no additional modifications made to the localization pipeline. Only graphs containing 256 nodes are considered for evaluation.

\autoref{tab:refinement_k_accuracy} presents the results of the evaluation. 
As we need to predict the position, like described in \autoref{sec:positionEstimation}, we need at least 3 neighbors. 
The findings reveal a clear trend: as the number of graphs involved in position estimation increases, accuracy decreases, and runtime correspondingly grows. Notably, for the experimental logistics hall floor at the Fraunhofer Institute, using three nearest embeddings or graphs proves optimal, achieving parity with the reference method.
However, the proposed localization pipeline performs position estimations more than 40 times faster than the reference, highlighting its significant potential for practical applications.

To evaluate the transferability of the localization pipeline to a different floor type, such as concrete, a best-out-of-10 approach was employed. This method analyzes the minimum pixel distance between a requested image and all other images within the dataset. Given that the framework requires a high degree of overlap between mapping images, the concrete dataset from Zhang et al. was selected due to its superior overlap compared to other datasets examined.

For this analysis, ten images were randomly selected from the dataset, while the remaining images were used to construct a leave-one-out Map-Database and Graph-Database for each selected image. 
This process resulted in ten distinct combinations of Map-Database and Graph-Database, which were thoroughly analyzed. 

From the ten trials conducted, the best result was selected as the representative score for the concrete floor type. 
This analysis yielded a best-out-of-ten pixel distance of \SI{181.9}{Pixel} for the concrete dataset. 
Notably, this pixel distance refers to the center of the image. 
When considering the image resolution of $1288 \times 964$ pixels; the result demonstrates the method's capability to adapt to diverse floor types, suggesting its potential for generalization beyond the specific conditions of the analyzed dataset.

\section{Conclusion and Future Work}
The results of our assessment highlight the exceptional potential of graph-based floor localization. This potential stems from its superior generalization capabilities and enhanced scalability, achieved by intentionally minimizing the search space. Our findings demonstrate that the proposed method attains nearly the same high accuracy as the reference implementation, detailed in \autoref{sec:eval}, while calculating a single position up to 40 times faster than the reference.

When using three nearest neighbors, the proposed approach achieves a position error of \SI{6.4}{mm}, a rotation error of \SI{0.3944}{\degree}, and a runtime of \SI{75.93}{ms}. These results underscore the significant efficiency and precision of the method.

Nevertheless, further research is required to refine this approach and fully meet the criteria for real-time robot localization. 
While scalability has been improved through the restructuring of the mapping process, additional investigations are necessary to optimize this aspect further. 
Moreover, evaluating larger datasets from the same and other floor types, combined with real ground truth data, is essential for thoroughly assessing the accuracy, scalability, and robustness of the method.

%
%
%
\bibliographystyle{splncs04}
%
\bibliography{bibliography}
\end{document}